\documentclass[11pt]{article}

\usepackage[preprint]{acl}

\usepackage{times}
\usepackage{latexsym}

\usepackage[T1]{fontenc}

\usepackage[utf8]{inputenc}

\usepackage{microtype}
\usepackage{booktabs}
\usepackage{tcolorbox}

\usepackage{inconsolata}

\usepackage{graphicx}

\title{Emergence of Context Characteristics Sensitivity in Large Language Models}

\def\authorsep{\hspace{0.3em}}
\author{\textbf{Nadya Yuki Wangsajaya\textsuperscript{1}} \authorsep \textbf{Haeun Yu\textsuperscript{2}} \authorsep \textbf{Isabelle Augenstein\textsuperscript{2}} \medskip\\
\null\textsuperscript{1}Nanyang Technological University \quad \null\textsuperscript{2}University of Copenhagen\\
\texttt{nady0006@e.ntu.edu.sg}}

\begin{document}
\maketitle
\begin{abstract}
During instruction fine-tuning (IFT), large language models (LLMs) learn to follow instructions by using the provided context to answer a query. While prior work has studied how context characteristics correlate with context usage by the LLM, this analysis has been limited to inference time, leaving open \emph{how} these relationships are acquired in the first place. Here, we measure how models' sensitivity to such characteristics shifts across successive IFT stages: supervised fine-tuning (SFT), direct preference optimization (DPO), and reinforcement learning with verifiable rewards (RLVR). Experiments across four models and three datasets show that SFT makes models more likely to use contexts that are easy to understand, such as containing high length, context-query similarity, and fluency. Post-SFT dynamics may either reinforce or resolve these preferences depending on the training dataset. Our findings reveal that context usage is actively reshaped at each IFT stage, and designing a balanced IFT dataset is important in ensuring robust context utilization of instruction-tuned models.\footnote{\url{https://github.com/copenlu/context-characteristics-sensitivity}}
\end{abstract}

\section{Introduction}
LLMs can use the given context to answer queries; a valuable trait as it allows models to access up-to-date knowledge without re-training \cite{mosbach-etal-2023-shot, ovadia-etal-2024-fine}. This enables various applications, from retrieval-augmented generation \cite{rag-one, gao2024retrievalaugmentedgenerationlargelanguage} to tool use and automated judges \cite{llm-judge, llm-judge-2}. However, models' context usage is often not robust; they sometimes ignore the provided context in favor of their parametric knowledge \cite{xie2024adaptive}, or are easily distracted by irrelevant context \cite{shi2023distracted}. 
The consequences can be significant: for example, inappropriate context usage may make models susceptible to injected misinformation \cite{pan-etal-2023-risk}. Understanding \emph{what} a model uses in its context is important because it ensures reliable model behavior.


\begin{figure}[t]
  \includegraphics[width= \columnwidth]{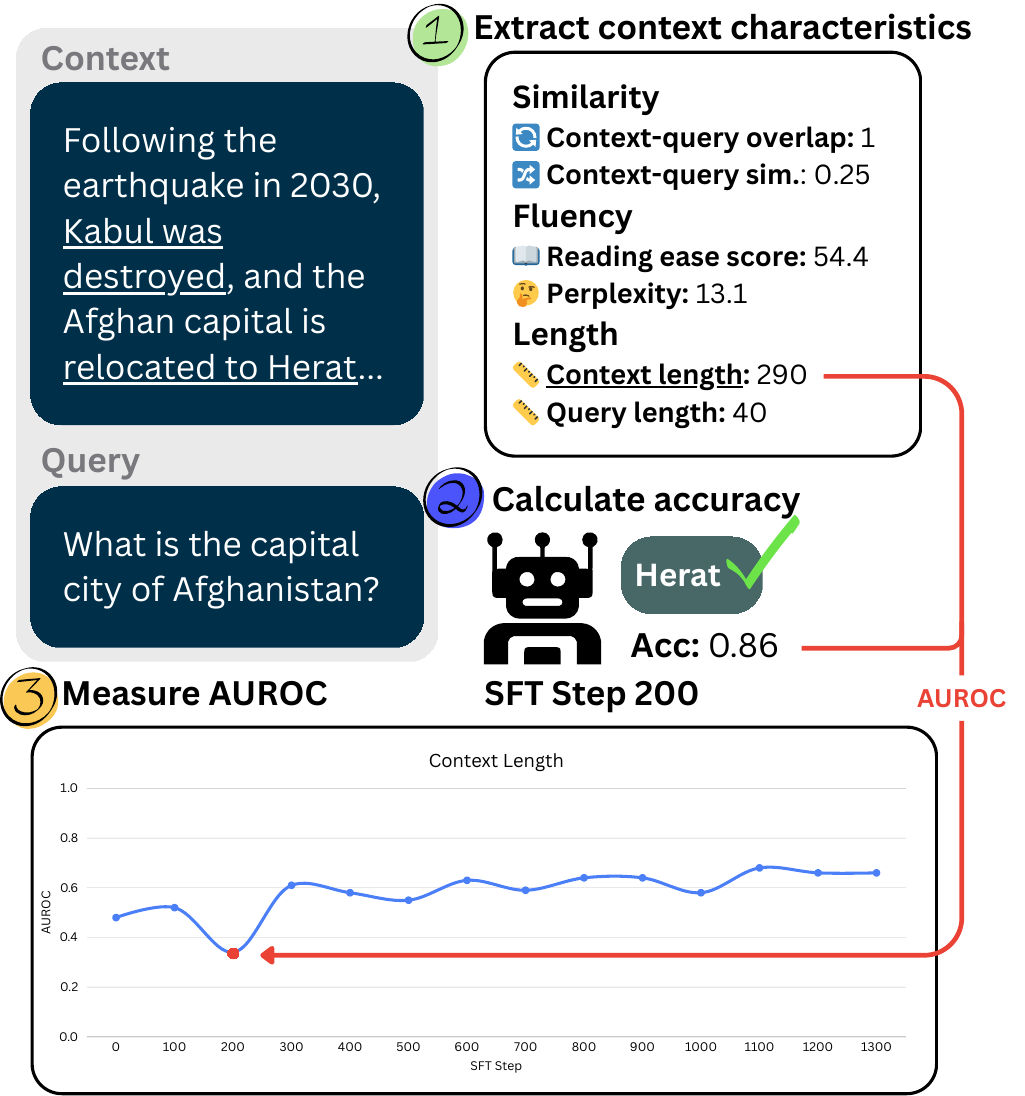}
  \centering
  \caption{\textbf{Experimental setup.} From each dataset, three families of context characteristics (similarity, fluency, and overlap) are extracted. Accuracy is then calculated based on the model's generated output. Sensitivity is measured by calculating the AUROC score (Section \ref{sec:results}) between each characteristic and accuracy.}
  \label{fig:experiment}
\end{figure}

An intuitive approach to understanding context usage is to investigate the characteristics of the context itself. Recent work studies lexical properties such as length, fluency, word overlap, and source reliability, finding that these characteristics correlate with context usage to varying degrees \cite{hagstrom-etal-2025-reality, hagstrom2026cubbenchmarkingcontextutilisation}. Yet, little is known about how the models learn to use context in the first place and, in particular, how sensitivities to certain context characteristics are acquired. In this work, \emph{context characteristics sensitivity} refers to the degree to which a characteristic's value influences whether the model uses the provided context to answer the query. Prior work has established that IFT plays a dominant role in teaching models to use context \cite{goyal2025contextparametric} and in shaping LLM behavior more broadly \cite{bhatia2025valuedriftstracingvalue}. By understanding how IFT training dynamics shape context characteristic sensitivity, we can curate datasets that elicit more robust context utilization in instruction-tuned LLMs.

Our analysis spans different IFT stages: SFT, DPO, and RLVR. Further, we train an LLM through SFT and DPO, and measure how this relationship evolves at each step; see Figure \ref{fig:experiment}. We find that context usage is actively reshaped across IFT stages. During SFT, LLMs develop a sensitivity towards easy-to-understand characteristics such as short and fluent contexts, and higher context-query similarity. The effect of DPO, in turn, is determined by the preference dataset, specifically the difference in lexical characteristics of the chosen and rejected responses. To ensure robust context usage and counteract the context characteristic sensitivities developed during SFT, the preference dataset must be carefully curated such that the chosen and rejected responses share the same context characteristics.

\section{Method}
\label{sec:method}
\subsection{Context Characteristics}
We examine three families of context characteristics aps by prior work on the influence of context characteristics on context usage: similarity, fluency, and length \cite{hagstrom-etal-2025-reality}. We drop the unreliability family as it is specific to fact-checking tasks. Implementation of the extraction process is detailed in Appendix \ref{app:extract}.

\paragraph{Similarity} LLMs often exploit high query-context similarity as a shortcut for whether to use context \cite{pmlr-v267-modarressi25a, shortcutlearning}, which we score with: \emph{Jaccard similarity} \cite{JaccardEtudeCD} after lowercasing and punctuation removal; and the fraction of named entities that appear both in the query and context (\emph{query-entity overlap}).

\paragraph{Fluency} LLMs tend to follow factually incorrect evidence when it is coherent and fluent \cite{xie2024adaptive, stureborg2024largelanguagemodelsinconsistent}. We measure fluency through \emph{Flesch reading ease score} \cite{fleschNewReadabilityYardstick1948}, a readability metric based on sentence length and syllable count, with higher scores indicating text that is easier to read, and \emph{Perplexity}, where lower values indicate the context is considered fluent under the model's learned distribution.

\paragraph{Length} Previous work documents two opposing forces affecting LLMs' context usage: they use long contexts less effectively \cite{xu2024retrieval}, but also treat longer passages as more credible \cite{dubois2024lengthcontrolled}.
We investigate this by measuring \emph{Context length} and \emph{Query length}, both calculated as the raw character counts of each string.

\subsection{Instruction Fine-Tuning}
\label{sec:ift}
We conduct (1) a stage-wise comparison of sensitivities across IFT stages, and (2) a step-wise analysis tracking how they evolve within each stage.

For the stage-wise analysis, we evaluate four models: Llama-3.1-8B, Llama-3.2-1B \cite{grattafiori2024Llama3herdmodels}, OLMo-2-1B, and OLMo-2-7B \cite{olmo20242olmo2furious}. We use the final IFT model (SFT, DPO, RLVR) of Llama-3.1-8B and OLMo 2 (1B and 7B variants) released by the Open Instruct library\footnote{\url{https://allenai.github.io/open-instruct/}\label{fn:open-instruct}} for our stage-wise analysis. Note that the Llama-3.1-8B model is trained using the Tulu3 dataset \cite{lambert2025tulu}, which is different from the datasets used to train the OLMo models detailed by \citet{olmo20242olmo2furious}.

Since the intermediate SFT and DPO checkpoints are not released, to conduct the step-wise analysis, we additionally fine-tune Llama-3.2-1B on the same SFT and DPO dataset as OLMo-2-1B. Llama-3.2-1B is chosen because it is small yet still able to use context well; see Appendix \ref{app:acc}. We adopt the hyperparameters of \citet{lambert2025tulu} unchanged. We exclude RLVR from step-wise analysis due to computational cost, but since it shares the same KL-regularized objective as DPO \cite{dpo-rl}, we expect DPO trends to extend to RLVR. Detailed IFT configuration is in Appendix \ref{app:hparams}.

\section{Experimental Setup}
\label{sec:setup}

\begin{figure*}[h]
    \includegraphics[width=\textwidth]{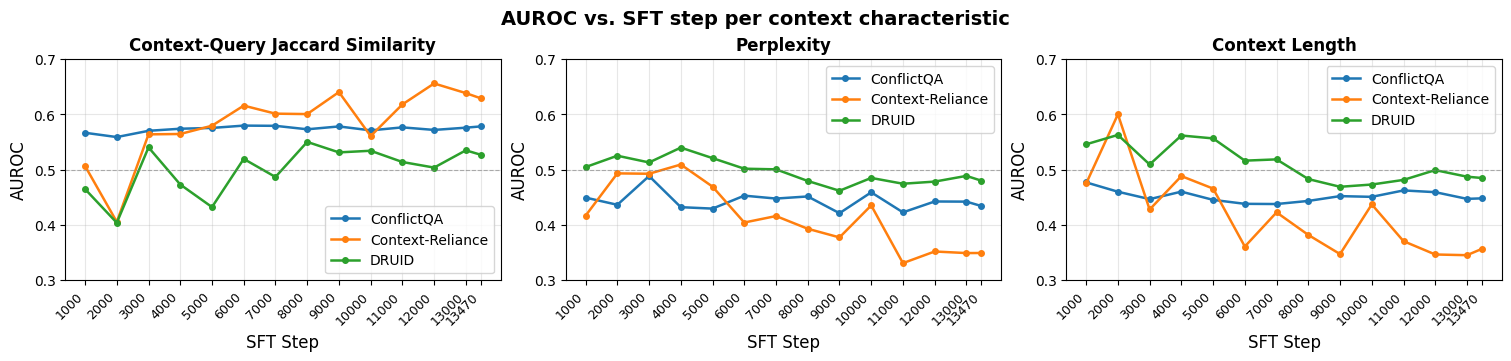}
    \includegraphics[width=\textwidth]{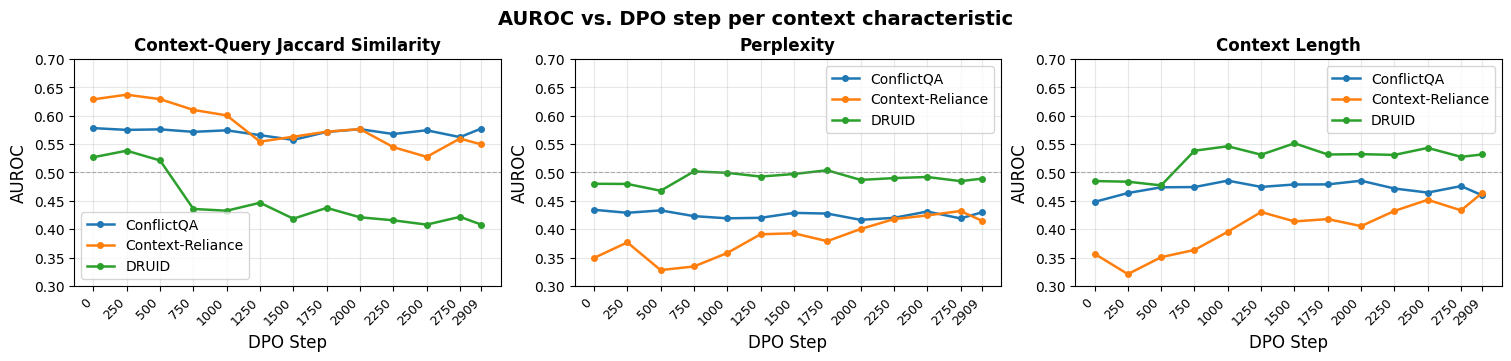}
    \caption{\textbf{Step-wise analysis across three datasets} of Llama-3.2-1B (Section \ref{sec:ift}). Here, we show one metric of each characteristic family; other metrics are in Appendix \ref{app:ccfull}. During SFT (top), the model develops sensitivities towards easy-to-understand context, while DPO (bottom) resolves the learned sensitivities.}
    \label{fig:step-wise}
\end{figure*}

\begin{figure*}[h]
    \includegraphics[width=\textwidth]{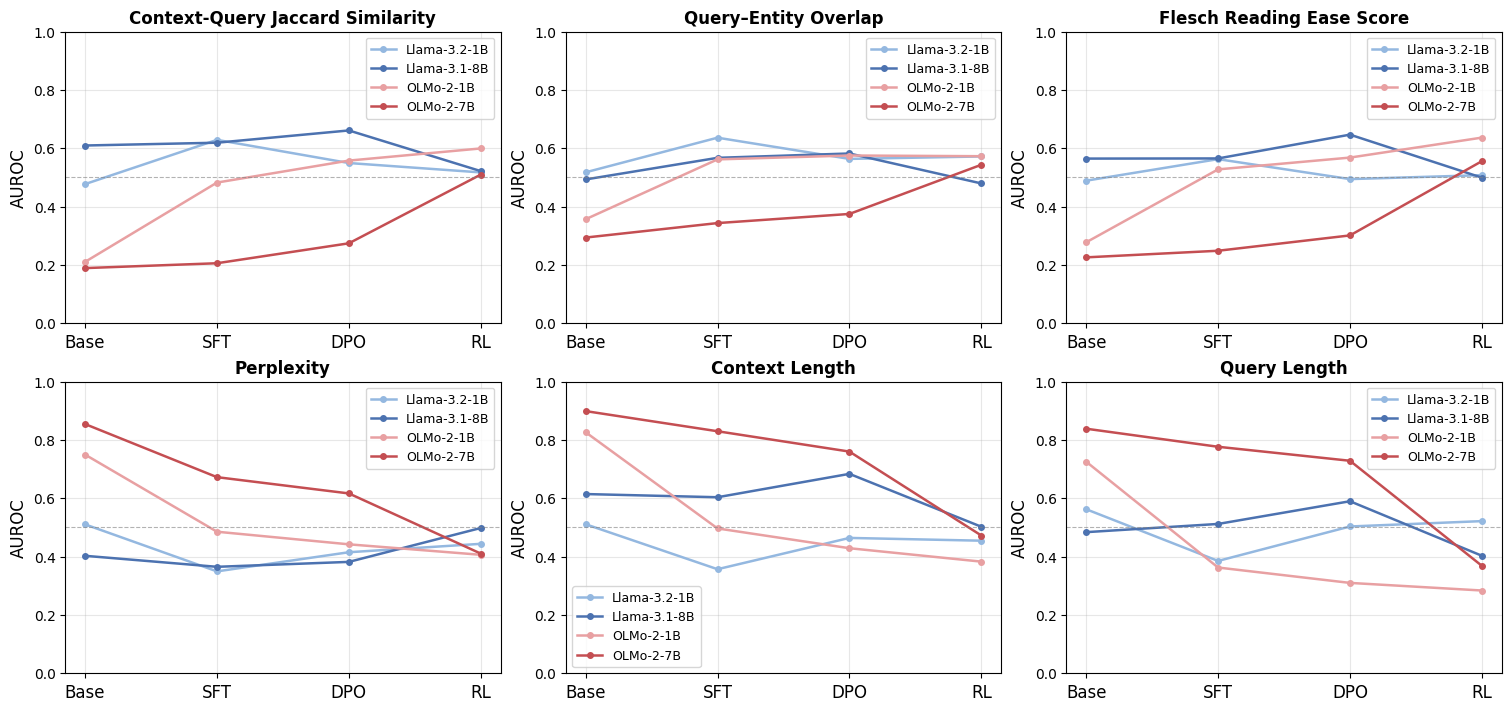}
    \caption{\textbf{Stage-wise analysis across four models on the Context-Reliance dataset}. SFT induces the same sensitivity in all models, while DPO's effect depends on characteristic differences in its training data. Evaluation results on ConflictQA and DRUID dataset are in Appendix \ref{app:allmodel}.} 
    \label{fig:stage-wise}
\end{figure*}

\paragraph{Datasets} 
We use three datasets chosen for diversity in task format and in the relationship between context and parametric knowledge. ConflictQA \cite{xie2024adaptive} and Context-Reliance \cite{goyal2025contextparametric} are open-ended QA tasks with counterfactual contexts, ensuring that accuracy purely reflects context use. In contrast, DRUID \cite{hagstrom-etal-2025-reality} is a claim verification task that reflects realistic context–knowledge interactions.

\paragraph{Evaluation}
For each combination of model, dataset, and characteristic, we compute the AUROC between the characteristic value and accuracy. The model's generation is considered accurate if it matches the context-based ground truth. We use AUROC because accuracy is binary, making other correlation measures, such as Pearson and Spearman, less appropriate. See Appendix \ref{app:setup} for details.

\section{Results}
\label{sec:results}
Figures \ref{fig:step-wise} and \ref{fig:stage-wise} show how context characteristics sensitivity evolves step-wise and stage-wise, respectively. To gauge sensitivity, we use the AUROC score, which measures how well the context characteristic ranks accurate examples above inaccurate ones, ranging from 0.5 (chance) to 1 (perfect separation). Values above 0.5 indicate high characteristic values correspond to accurate examples, while values below 0.5 suggest the inverse.

\begin{table*}[h]
\centering
\small
\setlength{\tabcolsep}{3pt}
\begin{tabular}{l ccc ccc}
\toprule
& \multicolumn{3}{c}{\textbf{Llama-3.2-1B}} & \multicolumn{3}{c}{\textbf{Llama-3.1-8B}} \\
\cmidrule(lr){2-4} \cmidrule(lr){5-7}
Characteristic & chosen & rejected & $\Delta$ & chosen & rejected & $\Delta$ \\
\midrule
Context-query Jaccard similarity & 0.158 & 0.150 & \textbf{0.008} & 0.192 & 0.161 & \textbf{0.031} \\
Query-entity overlap             & 0.409 & 0.376 & \textbf{0.033} & 0.395 & 0.354 & \textbf{0.041} \\
Flesch reading ease score        & 46.454 & 40.143 & \textbf{6.311} & 45.273 & -165.690 & \textbf{210.963} \\
Context length                   & 1844.392 & 1844.489 & \textbf{-0.097} & 1727.275 & 1705.265 & \textbf{22.010} \\
Query length                     & 946.300 & 946.300 & 0.000 & 910.274 & 910.274 & 0.000 \\
\bottomrule
\end{tabular}
\caption{\textbf{Mean context characteristics for chosen and rejected responses for the DPO dataset} used to train the Llama models in Figure \ref{fig:stage-wise}. $\Delta$ = chosen $-$ rejected. Bolded $\Delta$ indicates statistical significance ($p < 0.05$), more details in Appendix \ref{app:stats}.}
\label{tab:dpo-characterize}
\end{table*}

\subsection{Models learn sensitivities towards easy-to-understand contexts during SFT}
For each characteristic, Figures \ref{fig:step-wise} (top) and \ref{fig:stage-wise} show a consistent AUROC trend during SFT, across all models and all datasets. This indicates that models learn the same context characteristics sensitivities when deciding whether to follow the provided context.

Specifically, the AUROC score of \emph{Context-query Jaccard similarity} increases during SFT. Models increasingly use contexts that are lexically similar to the query. This is intuitive: higher token overlap signals relevance, making the context appear more directly applicable to the question. On the other hand, the AUROC score of \emph{Perplexity} decreases, meaning models increasingly follow low-perplexity contexts. Lower perplexity indicates more fluent, natural text, so the model learns to treat well-formed contexts as more trustworthy. Similarly, the AUROC score of \emph{Context length} decreases, suggesting models increasingly follow shorter contexts, possibly as they are easier to integrate compared to longer ones and therefore more reliable.

Our results support prior findings, where models are shown to use contexts with high similarity \cite{pmlr-v267-modarressi25a, shortcutlearning} and high fluency \cite{xie2024adaptive, stureborg2024largelanguagemodelsinconsistent} more frequently. Our investigation reveals that this sensitivity mainly forms during SFT. Notably, while context characteristic sensitivities increase across SFT steps, they do not translate into improved accuracy (Appendix \ref{app:acc}). The model exploits context characteristics as a shortcut; it uses context only when the context is easy-to-understand, rather than genuinely assessing its relevance.

\subsection{DPO sensitivities are shaped by characteristic differences in training data}
From Figure \ref{fig:step-wise} (bottom), DPO resolves the sensitivities learned during SFT by shifting sensitivities to the pre-SFT level. However, Figure \ref{fig:stage-wise} shows that Llama-3.1-8B does not follow the same trend during DPO. For example, DPO amplifies, rather than resolves, sensitivity towards \emph{Context-query Jaccard similarity} in the 8B variant. As the two models are trained on different DPO datasets (Section \ref{sec:ift}), we attribute this observation to the characteristics of the training dataset, specifically the difference between the chosen and rejected responses ($\Delta$) (Table \ref{tab:dpo-characterize}). We find that whether the post-DPO AUROC lies above or below 0.5 broadly tracks the sign of $\Delta$. This highlights the importance of balancing context characteristics of responses in the DPO dataset. Any imbalances can reinforce SFT-stage sensitivities and context usage that is not robust. We note a caveat: this pattern holds for QA datasets (ConflictQA and Context-Reliance), but is less consistent for DRUID (Figure \ref{fig:step-wise}; bottom), a claim verification task. This suggests that DPO's effect on context characteristic sensitivities may be task-dependent. Finally, we observe that RLVR mostly preserves the direction of DPO's effects as expected.

\section{Conclusion}
We investigate how the context characteristic sensitivity evolves across IFT stages. We find that SFT consistently instills the same sensitivities, favoring lexically similar, fluent, and shorter contexts, across datasets and model families. This sensitivity formation is distinct from accuracy, suggesting that SFT teaches models to exploit these easy-to-read characteristics to signal context usage, rather than robustly assessing the context. For DPO, we attribute sensitivity shifts directly to characteristic differences between chosen and rejected contexts, underscoring the need to design a balanced preference dataset to neutralize context characteristics sensitivities inherited from SFT. Together, these findings suggest that context usage is not a static model property but is actively shaped at each IFT stage, and can be steered through careful data curation choices.


\section*{Limitations}
Our study is limited to two model families (Llama and OLMo) at two scales each. We consider this scope appropriate for our research questions which focus on how training dynamics shape context characteristic sensitivity. However, broader coverage across different architectures remains an interesting open line of work. Additionally, we examine three families of context characteristics: similarity, fluency, and length. These are prevalent and well-grounded in prior work, though there may well be other characteristics that influence context usage, and would be worth exploring in future work.

An AI writing assistant tool is used for a grammar check and stylistic improvement. None of the AI generations is used as is.
 
\section*{Acknowledgments}
$\begin{array}{l}\includegraphics[width=1cm]{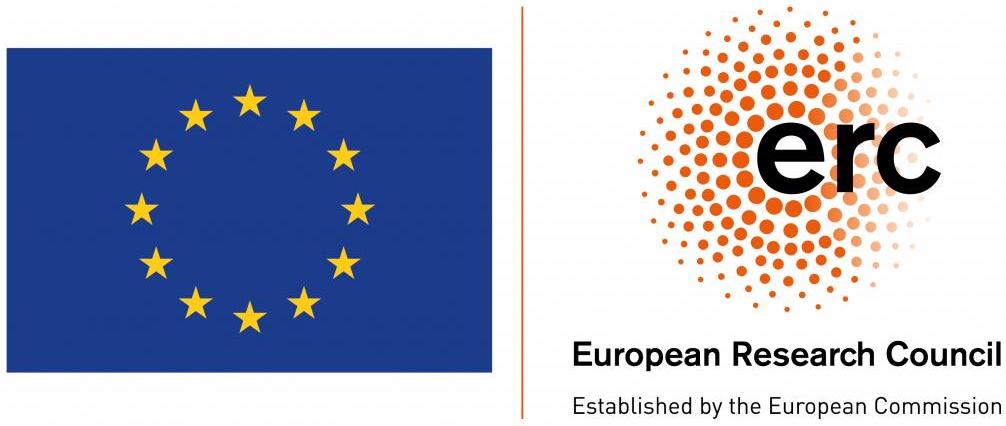} \end{array}$ 
This research was co-funded by the European Union (ERC, ExplainYourself, 101077481) and supported by the Pioneer Centre for AI, DNRF grant number P1. Views and opinions expressed are however those of the author(s) only and do not necessarily reflect those of the European Union or the European Research Council. Neither the European Union nor the granting authority can be held responsible for them. It was also funded by the College of Computing and Data Science at Nanyang Technological University and the CN Yang Scholars Programme. We also thank CopeNLU members at the University of Copenhagen for their careful proof-reading.

\bibliography{custom}

\appendix
\section{Context Characteristic Extraction}
\label{app:extract}
\paragraph{Similarity} \emph{Context-query similarity} is computed as the Jaccard index between the sets of lowercased tokens of the context and the query, after tokenization with NLTK \cite{loper2002nltknaturallanguagetoolkit} and removal of punctuation and typographic quote characters. \emph{Query-entity overlap} is computed by extracting named entities from both the context and the query using \texttt{spaCy}'s transformer-based pipeline \cite{Honnibal_spaCy_Industrial-strength_Natural_2020} and reporting the fraction of query entities that also appear in the context; queries with no detected entities are excluded.

\paragraph{Fluency} \emph{Flesch reading ease score} is calculated via the \texttt{textstat} library\footnote{\url{https://github.com/textstat/textstat}}. Meanwhile, \emph{Perplexity} is calculated according to a HuggingFace guide\footnote{\url{https://huggingface.co/docs/transformers/perplexity}}.

\paragraph{Length} \emph{Context length} and \emph{Query length} are measured as the raw number of characters in the corresponding strings.

\section{Llama-3.2-1B Accuracy}
\label{app:acc}
\begin{figure}[h]
    \centering
    \includegraphics[width=\columnwidth]{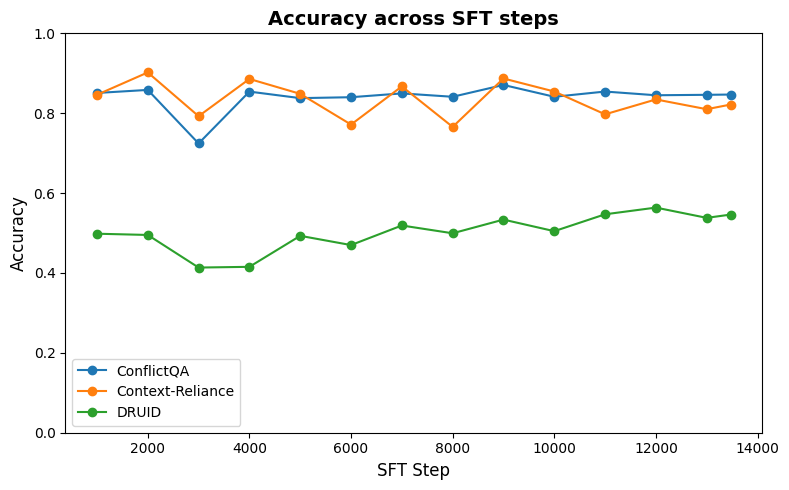}
    \includegraphics[width=\columnwidth]{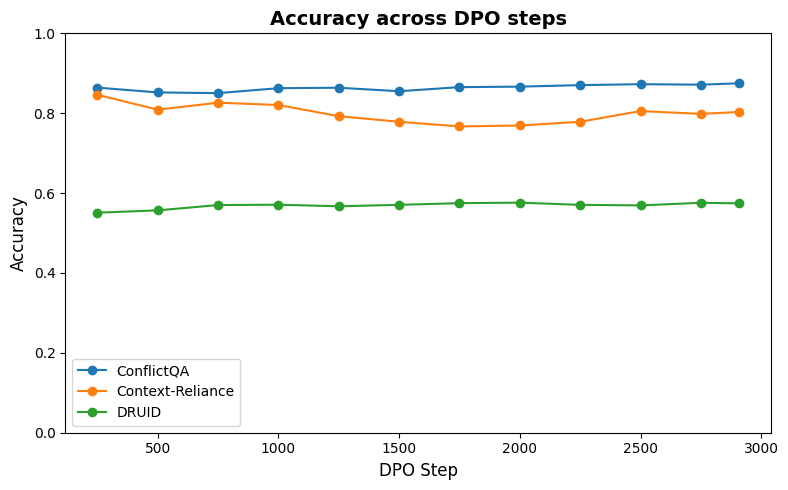}
    \caption{\textbf{Context-based accuracy of Llama-3.2-1B per step}, throughout SFT (top) and DPO (bottom).}
    \label{fig:acc_combined}
\end{figure}

Figure \ref{fig:acc_combined} shows that small Llama-3.2-1B model are able to use context properly. For ConflictQA and Context-Reliance, the accuracy throughout SFT and DPO training hovers around 0.8 to 0.9. For DRUID, accuracy is lower at around 0.5, but it is still higher than random chance (0.33; three-way classification). This shows that we can use Llama-3.2.1B as a proxy to study changes in context usage during SFT and DPO.

\section{Configuration for Instruction Fine-Tuning}
\label{app:hparams}
Table \ref{tab:training-configs} shows the config for IFT used to train Llama-3.2-1B, which checkpoints are used to create Figure \ref{fig:step-wise}.

\begin{table}[h]
\centering
\small
\begin{tabular}{l l l}
\toprule
\textbf{Stage} & \textbf{Parameter} & \textbf{Value} \\
\midrule
SFT & Learning rate               & 3e-5 \\
                     & Batch size/device           & 2 \\
                     & Gradient accumulation steps & 16 \\
                     & Epochs                      & 2 \\
                     & Max sequence length         & 4096 \\
                     & LR scheduler                & Linear \\
\midrule
DPO & Learning rate               & 2.5e-6 \\
                     & Batch size/device           & 2 \\
                     & Gradient accumulation steps & 16 \\
                     & Epochs                      & 1 \\
                     & Max sequence length         & 2048 \\
                     & LR scheduler                & Linear \\
                     & $\beta$                     & 5 \\
\bottomrule
\end{tabular}
\caption{Detailed configuration for SFT and DPO fine-tuning.}
\label{tab:training-configs}
\end{table}

We save intermediate checkpoints every 1,000 steps during SFT and every 250 steps during DPO. All fine-tuning is done on 4XA100 GPU. SFT took 24 hours and DPO took 6 hours. Conducting RLVR would have taken approximately 344 hours (14X more compute than SFT).

\section{Full Context-Characteristic Results}
\label{app:ccfull}
\begin{figure*}[h]
    \includegraphics[width=\textwidth]{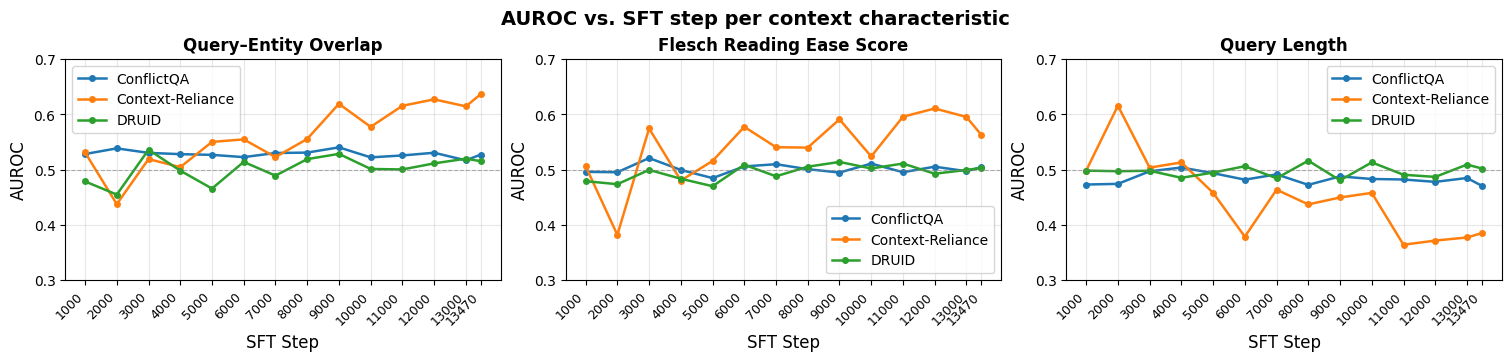}
    \includegraphics[width=\textwidth]{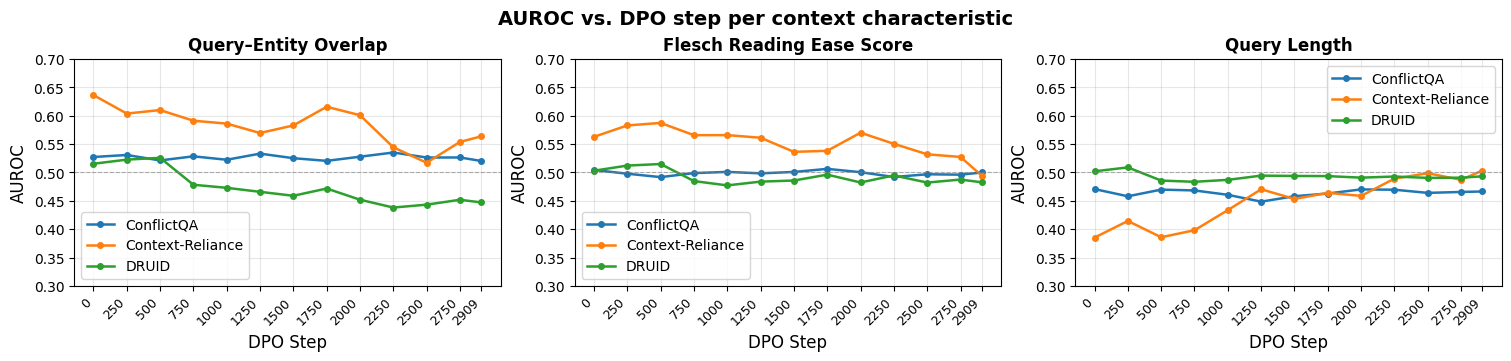}
    \caption{\textbf{Step-wise analysis between the remaining other context characteristic with accuracy} including the unreliable context characteristic family. Llama-3.2-1B is used (Section \ref{sec:ift}).}
    \label{fig:step-wise-others}
\end{figure*}

The step-wise relationship for \emph{Query-entity overlap}, \emph{Flesch reading ease score}, and \emph{Query length} is shown in Figure \ref{fig:step-wise-others}.

\section{Detailed Set-up}
\label{app:setup}
For DRUID (5,490 rows), predictions are mapped to one of the three labels (\textit{supports}, \textit{refutes}, \textit{insufficient}) via string matching against the generated output; mismatches and abstentions count as incorrect. For ConflictQA (8,067 rows) and Context-Reliance (1,064 rows), we extract the context-based and parametric-based ground truths and string-match the model's output against them. When the result is ambiguous (both or neither match), we fall back to an LLM-as-a-judge using Qwen-3.6 Plus \cite{qwen36plus}, as it performed best in held-out evaluation set (See Appendix \ref{app:judge}). Prompt used for inference are shown in Appendix \ref{app:prompts}.

All inference uses greedy decoding with \texttt{max\_new\_tokens=1024} and a fixed random seed (1000) across PyTorch, CUDA, NumPy, and Python; cuDNN is set to deterministic mode to ensure reproducibility. Inference is run on a single A100 GPU.

\section{Extended Model-Dataset Results}
\label{app:allmodel}
\begin{figure*}[h]
    \includegraphics[width=\textwidth]{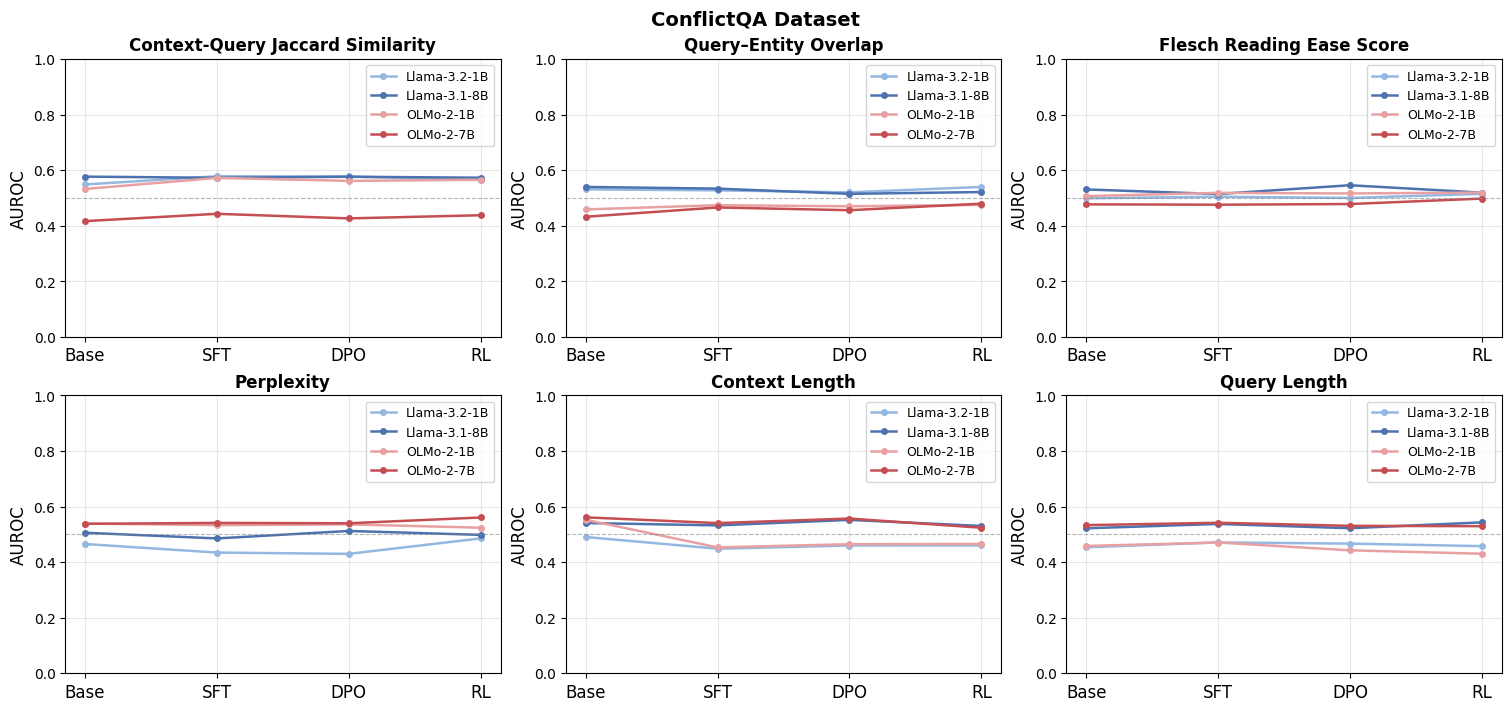}
    \includegraphics[width=\textwidth]{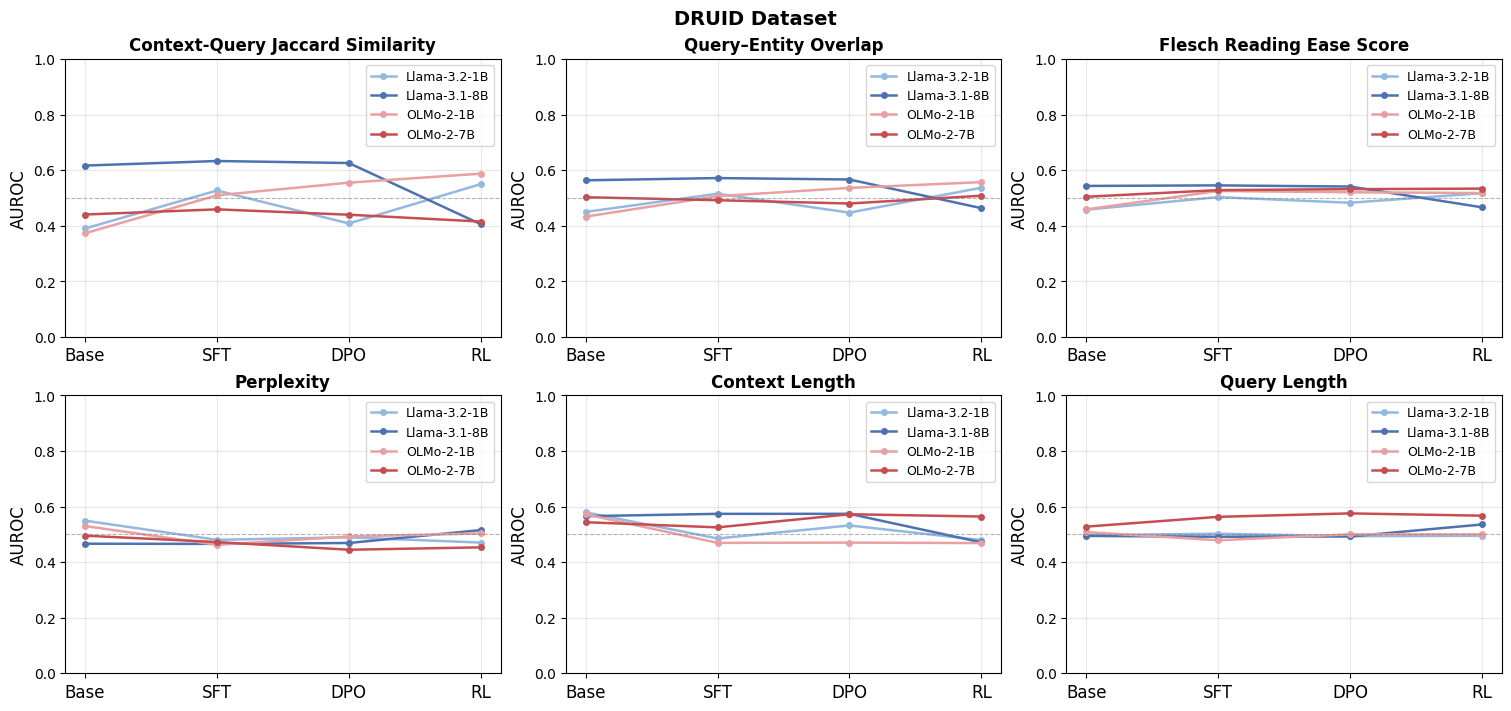}
    \caption{\textbf{Stage-wise analysis across four models on two extra datasets}: ConflictQA (top) and DRUID (bottom).}
    \label{fig:stage-wise-all}
\end{figure*}

Figure \ref{fig:stage-wise-all} shows the stage-wise relationship using ConflictQA and DRUID datasets. 

\section{Statistical Tests}
\label{app:stats}
We use Wilcoxon signed rank test to determine significance. It tests the null hypothesis that two related paired samples (the chosen and rejected answers from the DPO training dataset) come from the same distribution. Since $p<0.05$ for all of $\Delta$ in Table \ref{tab:dpo-characterize}, the null hypothesis is rejected. The difference ($\Delta$) is significant.

Student t-test is not used as we are unable to assume the difference in mean of each context characteristic is normally distributed.

\section{LLM Judge}
\label{app:judge}
For ConflictQA and Context-Reliance dataset, if the model's answer contains both (or neither of) context-based and parametric-based ground truths, we fall back to LLM judge to determine accuracy. Figure \ref{fig:prompt-judge} shows the prompt for the LLM judge.

To find the best judge model to use, we manually annotate 50 samples of model outputs (using the same prompt in Figure \ref{fig:prompt-judge}) and validate them with 10 different judges. Agreement between each judge and the manual annotation is shown in Table \ref{tab:judge-agreement}. For our analysis, we use Qwen-3.6 Plus as it is the best-performing and most cost-efficient judge. All judge inferences use the OpenRouter \footnote{\url{https://openrouter.ai/}} API. Only input pricing is considered, as the model's output is short and therefore negligible.

\begin{figure}[h]
\begin{tcolorbox}[
  colback=white,
  colframe=red!70!black,
  coltitle=white,
  colbacktitle=red!70!black,
  title={\textbf{Prompt for LLM Judge}},
  boxrule=1pt,
  sharp corners
]
You are a judge.\\
Your task: determine whether a generated answer correctly matches a reference answer.\\ \\
Rules: \\
- Output ONLY the digit 1 or 0. Nothing else.\\
- Output 1 if the generated answer is correct, i.e., it expresses, implies, or contains the reference answer.\\
  - Partial matches are acceptable (e.g., "Trump" matches "Donald Trump"), but not wrong answers (e.g., "Hudson Williams" does not match "Hudson Lee").\\
  - Paraphrases are acceptable (e.g., "What is the capital of France? It is moved from Paris to Versailles" matches "Versailles").\\
- Output 0 if the generated answer does not answer the question directly, or require strong inference to get the answer.\\
- Output 0 if the generated answer is wrong, irrelevant, or contradicts the reference answer.\\
- If there is an explicit final answer in the generated answer, only consider the final answer in your decision.\\ \\
Question: "\textless{}question\textgreater{}"\\
Reference answer: "\textless{}context-based ground truth\textgreater{}"\\
Generated answer: "\textless{}model's output\textgreater{}"\\ \\
Does the generated answer match the reference answer? Output 1 or 0.
\end{tcolorbox}

\caption{Prompt used for LLM judge as a tie-breaker.}
\label{fig:prompt-judge}
\end{figure}

\begin{table}[h]
\centering
\small
\begin{tabular}{l l r}
\toprule
\textbf{LLM Judge} & \textbf{$\kappa$} & \textbf{Input Price} \\
\midrule
Gemini 2.5 Flash             & 0.716 & 0.30 \\
Qwen3 235B                   & 0.676 & 0.07 \\
Qwen 3.6 Plus                & \textbf{0.920} & 0.17 \\
Llama 4 Maverick             & 0.735 & 0.15 \\
Llama 3.3 70B                & 0.401 & 0.10 \\
MiniMax M2.7                 & 0.671 & 0.28 \\
GLM-5.1                      & 0.880 & 0.98 \\
GLM-4.7                      & \textbf{0.920} & 0.45 \\
GPT 5.4-mini (high reasoning) & \textbf{0.920} & 0.75 \\
GPT 5.4-mini (low reasoning)  & 0.880 & 0.75 \\
\bottomrule
\end{tabular}
\caption{LLM Judge agreement (Cohen's $\kappa$) with manual annotation and API input pricing (US\$/M tokens).}
\label{tab:judge-agreement}
\end{table}

\section{Prompts Used for Inference}
\label{app:prompts}
Figure \ref{fig:prompts} shows the prompts used for inference, based on the task.
\begin{figure}[h]
\begin{tcolorbox}[
  colback=white,
  colframe=red!70!black,
  coltitle=white,
  colbacktitle=red!70!black,
  title={\textbf{Prompt for Claim Verification Task (DRUID)}},
  boxrule=1pt,
  sharp corners
]
Based on the provided context, is the claim True or False? If you are not sure or cannot answer, say None.\\
\textless{}user\textgreater{}\\
Context: "\textless{}context\textgreater{}"\\
Claim: "\textless{}claim\textgreater{}"\\
\textless{}assistant\textgreater{}
\end{tcolorbox}

\begin{tcolorbox}[
  colback=white,
  colframe=red!70!black,
  coltitle=white,
  colbacktitle=red!70!black,
  title={\textbf{Prompt for Open-ended QA Task (ConflictQA, Context-Reliance)}},
  boxrule=1pt,
  sharp corners
]
Answer the following reading comprehension question.\\
\textless{}user\textgreater{}\\
Context: "\textless{}context\textgreater{}"\\
Query: "\textless{}query\textgreater{}"\\
\textless{}assistant\textgreater{}
\end{tcolorbox}

\caption{Prompts used for model output generation.}
\label{fig:prompts}
\end{figure}

\end{document}